\documentclass[10pt,twocolumn,letterpaper]{article}

\usepackage{iccv}   
\usepackage{times}
\usepackage{graphicx}
\usepackage{amsmath}
\usepackage{amssymb}
\usepackage{algorithm}
\usepackage{algorithmicx}
\usepackage{multirow}
\usepackage[american]{babel}
\usepackage{pifont}
\usepackage{subcaption}
\newcommand{\cmark}{\text{\ding{51}}}
\newcommand{\xmark}{\text{\ding{55}}}
\graphicspath{{figure/}}

\usepackage[pagebackref=true,breaklinks=true,colorlinks,bookmarks=false]{hyperref}

\def\iccvPaperID{3232} % *** Enter the ICCV Paper ID here

\iccvfinalcopy
\newcommand{\squishlist}{
	\begin{list}{$\bullet$}
		{ \setlength{\itemsep}{0pt}
			\setlength{\parsep}{1pt}
			\setlength{\topsep}{1pt}
			\setlength{\partopsep}{0pt}
			\setlength{\leftmargin}{1.em}
			\setlength{\labelwidth}{1em}
			\setlength{\labelsep}{0.5em} } }
\newcommand{\squishend}{\end{list} 
}
\ificcvfinal\fi
\begin{document}

\title{Weakly Supervised Foreground Learning for\\Weakly Supervised Localization and Detection }
\author{Chenlin Zhang\\
Nanjing University\\
{\tt\small zhangcl@lamda.nju.edu.cn}
\and
Yin Li\\
University of Wisconsin-Madison\\
{\tt\small yin.li@wisc.edu}
\and
Jianxin Wu\\
Nanjing University\\
{\tt\small wujx2001@nju.edu.cn}
}
\maketitle

\begin{abstract}
 Modern deep learning models require large amounts of accurately annotated data, which is often difficult to satisfy. Hence, weakly supervised tasks, including weakly supervised object localization~(WSOL) and detection~(WSOD), have recently received attention in the computer vision community. In this paper, we motivate and propose the weakly supervised foreground learning (WSFL) task by showing that both WSOL and WSOD can be greatly improved if groundtruth foreground masks are available. More importantly, we propose a complete WSFL pipeline with low computational cost, which generates pseudo boxes, learns foreground masks, and does not need any localization annotations. With the help of foreground masks predicted by our WSFL model, we achieve 72.97\% correct localization accuracy on CUB for WSOL, and 55.7\% mean average precision on VOC07 for WSOD, thereby establish new state-of-the-art for both tasks. Our WSFL model also shows excellent transfer ability.
\end{abstract}

\section{Introduction}

Deep learning have been the \emph{de facto} standard for many tasks in the computer vision community. However, the thirst for large-scale labeled data has grown with the development of such models. Since accurate annotations are expensive and sometimes even unavailable, weakly supervised methods, including both weakly supervised object localization~(WSOL) and weakly supervised object detection~(WSOD) are popular these days. WSOL aims at predicting objects’ locations in test images when only the class-level labels of training images are given. Moving beyond WSOL, WSOD seeks to detect and classify multiple objects, which is considerably more challenging.

A recent work~\cite{evaluatecvpr2020} argues that previous WSOL methods in fact do not outperform the pioneering class activation map~(CAM)~\cite{camcvpr2016} method, and further claims that WSOL is an ill-posed problem when it is not given any location annotations. In~\cite{evaluatecvpr2020}, using only a few images with groundtruth pixel-level annotation (\ie, few-shot) can outperform existing WSOL methods. This few-shot learning setting, however, is not weakly supervised anymore.
\begin{figure}[t]
	\centering
		\includegraphics[width=0.6\linewidth]{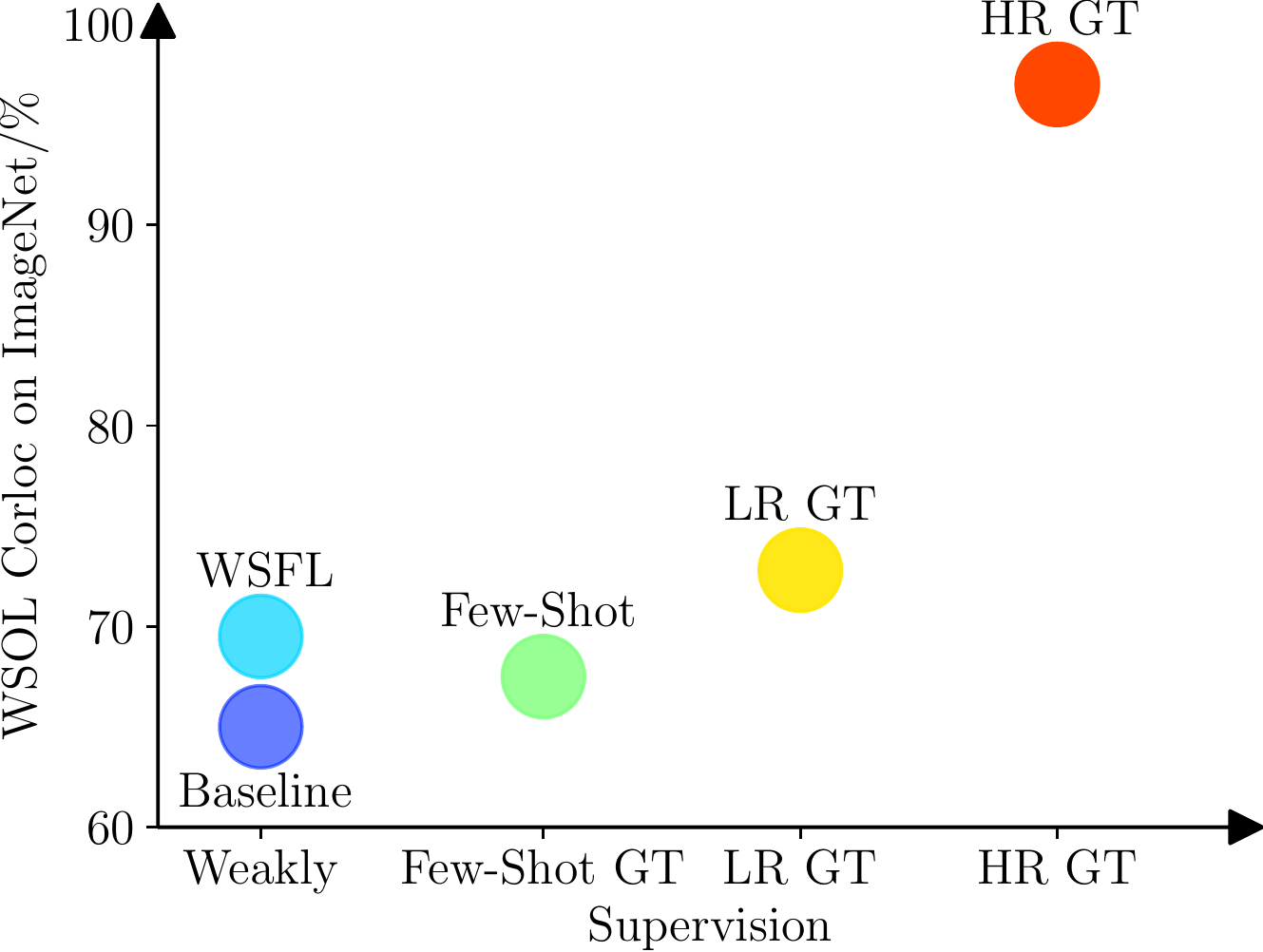}
	\vspace{-6pt}
	\caption{WSOL performance with different levels of supervision on ImageNet.}
	\label{fig:wsol banner}
	\vspace{-1em}
\end{figure}
\begin{figure}
	\centering
		\includegraphics[width=0.6\linewidth]{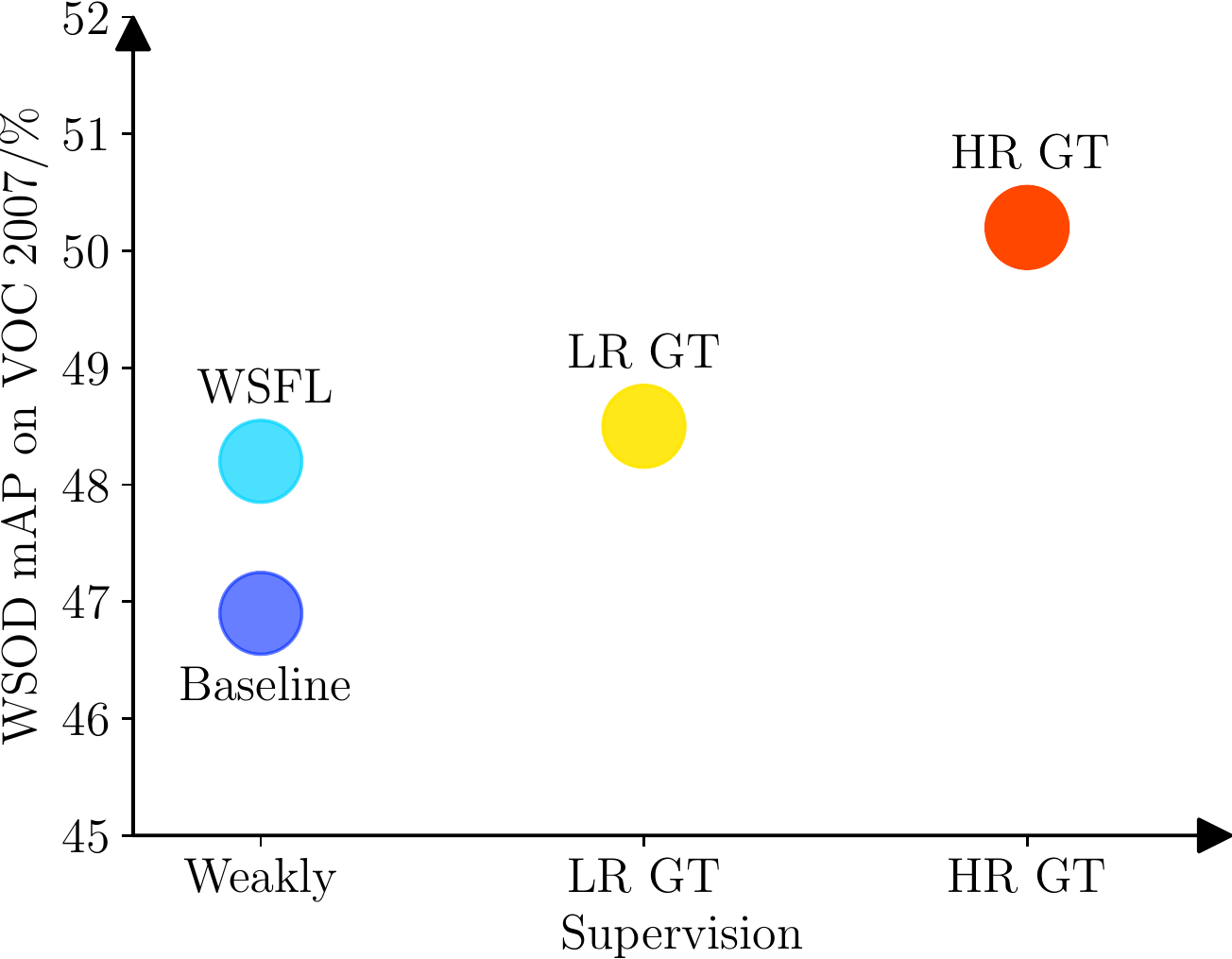}
	\vspace{-6pt}
	\caption{WSOD performance with different levels of supervision on VOC2007.}
	\label{fig:wsod banner}
	\vspace{-8pt}

\end{figure}
Although WSOL might be ill-posed inside the CAM framework, we propose a new task of \emph{weakly supervised foreground learning} (WSFL) and argue that it is feasible and well-posed. That is, \emph{it is viable to determine whether a pixel belongs to foreground (object) or background with only weak supervision.} As the method PSOL~\cite{psolcvpr2020} illustrated, in WSOL we need to separate two subtasks (localization of a foreground object and recognition of its label), although existing methods mostly mix both subtasks together. The separated localization subtask in fact hinges on the WSFL task we propose, and the success of PSOL~\cite{psolcvpr2020} in turn gives us some confidence in the feasibility of WSFL.

The value of foreground/background dichotomy (and thus the proposed WSFL task) is beyond WSOL. Figures~\ref{fig:wsol banner} and~\ref{fig:wsod banner} illustrate that it is also very valuable in the more difficult WSOD setup. The dichotomy can appear in different forms with the supervision signal getting weaker and weaker (only provided for training images, more details in Sec.~4):
\squishlist
	\item High-resolution (HR) GT: Every pixel has a groundtruth (GT) foreground mask;
	\item Low-resolution (LR) GT: GT masks exist, but are low resolution (\eg, $14 \times 14$ mask for $224 \times 224$ image);
	\item High-resolution few-shot~\cite{evaluatecvpr2020}: Like HR GT, but only labeled few training images per category;
	\item WSFL: The proposed method, where masks are learned in a weakly supervised manner (\ie, no additional supervision);
	\item No GT (Baseline): No masks are used at all.
\squishend
Both figures show that extra supervision (masks) are extremely useful, no matter it is high- or low-resolution, or many- or few-shot. A natural question is: can we perform weakly supervised foreground learning (WSFL)? If the answer is ``yes’’,  will WSFL be on par with relatively weak masks (such as few-shot GT or low-resolution GT)?

\begin{figure*}[t]
	\centering
		\includegraphics[width=0.8\linewidth]{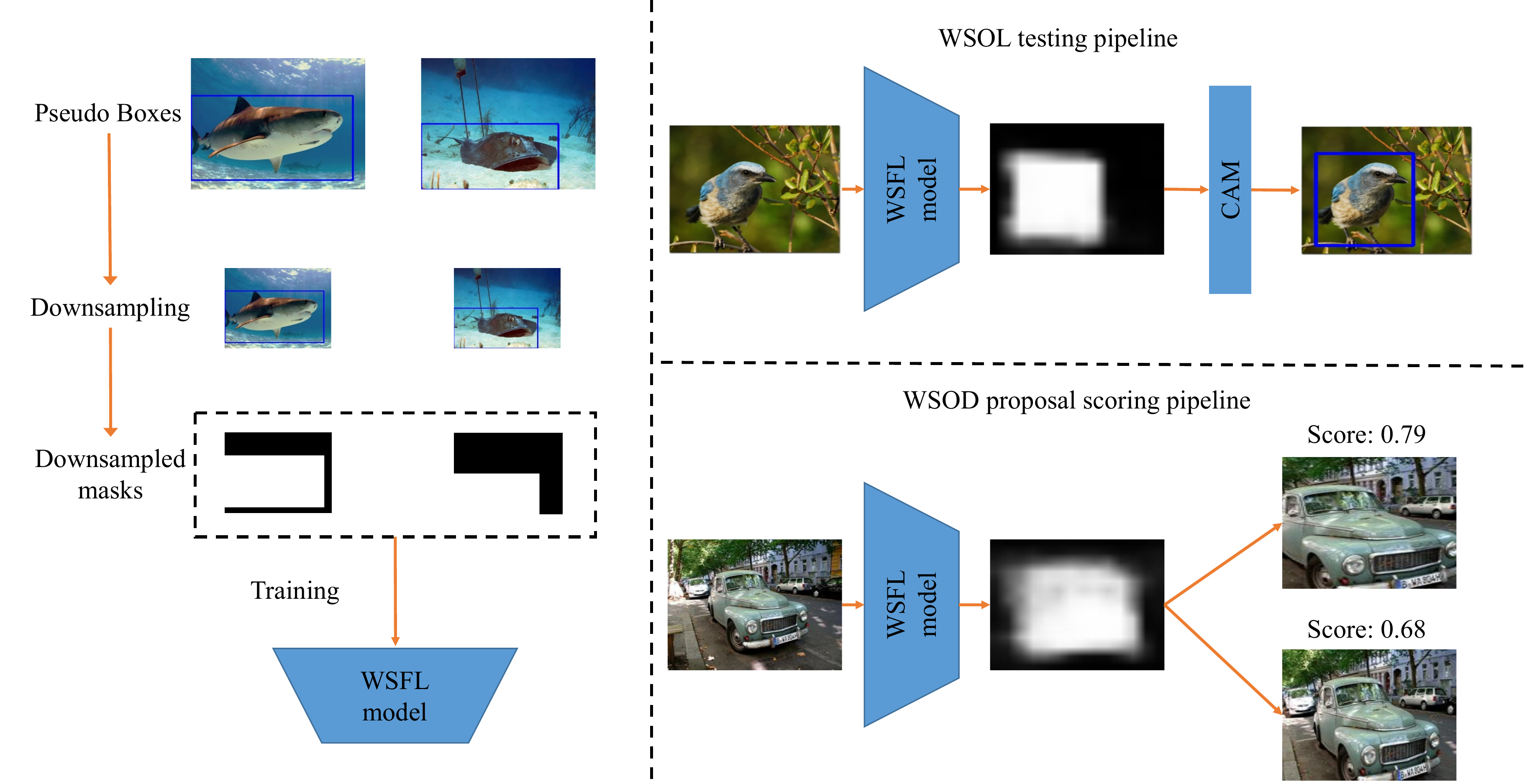}
	\vspace{-6pt}
	\caption{Overall pipeline of our WSFL framework. The training pipeline of WSFL lies on the left side while the testing pipeline of WSFL, including downstream applications WSOL and WSOD, lies on the right side. In WSOL, foreground masks predicted by WSFL is post-processed by CAM to generate bounding boxes. In WSOD, the foreground masks are used to filter away proposals with low foreground (objectness) scores. Both applications can be significantly improved by WSFL.}
	\label{fig:overall framework}
	\vspace{-6pt}
\end{figure*}

We believe that the answers to both questions are ``yes’’ and propose a WSFL pipeline, whose effectiveness is verified on various WSOL and WSOD datasets. The pipeline is illustrated in Fig.~\ref{fig:overall framework}. WSFL first generates pseudo boxes with a co-localization method, deep descriptor transformation~\cite{ddtpr2019}. Then, it generates low resolution pseudo foreground masks using the pseudo bounding boxes. A low resolution pixel mask classification model will be learned using the generated foreground masks. Compared to existing saliency methods~\cite{saliencyref1iccv2019,saliencyref2iccv2019,saliencyref3cvpr2019,saliencylataaai2018} that will also output a binary pixel mask using segmentation-based backbone models, our WSFL pipeline is computationally more efficient with classification backbone models.

During testing, WSFL directly predicts low resolution foreground masks using the learned low resolution pixel mask classification model. These masks can be combined with class activation map~(CAM) to predict bounding boxes in WSOL tasks. WSOD methods can also make use of these masks as extra supervision for objectness scores during training. With the proposed WSFL models, we achieve new state-of-the-art results for both WSOL and WSOD.

In short, our contributions are:
\squishlist
    \item We find that groundtruth foreground masks can greatly benefit tasks such as WSOL and WSOD.
    \item We propose WSFL (weakly supervised foreground learning) which learns foreground masks in a weakly supervised fashion (\ie, no extra supervision).
    \item Applying WSFL to WSOL and WSOD, our method establishes new state-of-the-art results for both tasks.
\squishend

\section{Related Works}

We briefly review recent works on weakly supervised object localization/detection, and saliency related methods.

\textbf{WSOL:} Weakly supervised object localization~(WSOL) aims at locating an object when given only image-level labels during training. Researchers have tried to adopt deep learning models to WSOL. The pioneering work~\cite{camcvpr2016} proposes class activation map~(CAM), which uses a combination of classification weights and feature map of a convolutional neural network~(CNN) to conduct localization.

Some methods improve the CAM pipeline~\cite{gradcamiccv2017,rethinkcameccv2020}.~\cite{evaluatecvpr2020} builds a new evaluation pipeline to deal with unfair comparisons. In addition, \cite{evaluatecvpr2020} also proposes a simple few-shot baseline, which uses a fully convolutional network~(FCN) model~\cite{fcncvpr2015}, and beats previous WSOL models. PSOL~\cite{psolcvpr2020} shows that the localization subtask should be separated from the recognition subtask, and proposes an independent localization model.

Another track of WSOL attack is to improve the localization ability of classification models.  ACoL~\cite{acolcvpr2018} conducts adversarial learning over baseline models. SPG~\cite{spgeccv2018} adopts self-produced localization masks into the base network. ADL~\cite{adlcvpr2019} combines attention dropout layer into the classification model. HaS~\cite{hideandseekiccv2017} and EIL~\cite{eilcvpr2020} use erasing to boost the localization performance. I$^2$C~\cite{i2ceccv2020} explores the inter-image information for localization in WSOL.  

\textbf{WSOD:} Weakly supervised object detection~(WSOD) seeks to detect multiple objects in a test image given only the image-level labels during training. Compared to WSOL, WSOD is more challenging. WSOD methods often use object proposals as extra inputs to the detection model. 

Most WSOD methods use the multi-instance learning framework to build a detector. Some works use external information like object size~\cite{wsodsizeeccv2016} or context information~\cite{contextlocneteccv2016}. The pioneering WSDDN method~\cite{wsddncvpr2016} uses spatial and class regularization over different proposals. OICR~\cite{oicrcvpr2017} introduces online instance refinement and pseudo groundtruth mining. The OICR framework is popular among WSOD works. One trend is to improve the pseudo groundtruth mining part~\cite{pcltpami2018,cmilcvpr2019}. Recently, MIST~\cite{wetectroncvpr2020} improves the mining rule of pseudo groundtruth boxes and proposes the Concrete DropBlock module.~\cite{enableresneteccv2020} makes ResNet~\cite{resnetcvpr2016} backbones work in WSOD tasks.

Extra information can guide the detector's learning process. WS-JDS~\cite{wsjdscvpr2019}, SDCN~\cite{sdcniccv2019} and OCRepr~\cite{ocreprmm2020} build a joint detection-segmentation framework and improves WSOD. WSOD$^2$~\cite{wsod2iccv2019} uses low-level vision information to help classify foreground proposals in the image.

\textbf{Saliency Methods:} Saliency aims at locating visually salient objects~\cite{saliencytpami2021}. Similar to our WSFL, saliency methods will also produce a binary mask for an input image, and weakly supervised saliency~(WSS) methods are closely related to this paper. \cite{wsscvpr2017} first uses deep CNNs in WSS. LICNN~\cite{saliencylataaai2018} formulates lateral inhibition to conduct WSS detection. SuperVAE~\cite{saliencyvaeaaai2019} introduces variational autoencoder into the field. However, WSS methods have high computational costs since they are based on segmentation models, and can only be trained on small datasets around 10,000 images. In contrast, our WSFL model can be applied in large-scale datasets like ImageNet-1k. 

\section{Weakly Supervised Foreground Learning}

We now present our WSFL framework, and describe how WSFL can be used in WSOL and WSOD.

\subsection{Notation}
Suppose we have a dataset $D$ comprising of $n$ images $\{I_{1},\ldots,I_{n}\}$, where each $I_i \in \mathbb{R}^{H \times W}$ is an image with height $H$ and width $W$. Each image $I_i$ has an object label $L_{i}$. In the fully-supervised setting, it also has a set of $m_i$ bounding box annotations $b_1,\ldots,b_{m_i}$, where $b$ is in the format of $(x_1,x_2,y_1,y_2)$, with $(x_1,y_1)$ and $(x_2,y_2)$ being the top-left and bottom-right corners of the bounding box. These bounding boxes are, however, not available in our weakly supervised setting. Feeding an image $I$ into a CNN model $W$, we will obtain a final feature map $F$ before proceeding to the classification head: $ F \in \mathbb{R}^{h \times w \times d}$ , with $h$,$w$ and $d$ being the height, width and depth of the feature map. 

\subsection{Motivating Weakly Supervised Foreground Learning}

For WSOL, an FCN (fully convolutional network~\cite{fcncvpr2015}) style network is used in~\cite{evaluatecvpr2020} to train a binary segmentation network that predicts per-pixel foreground masks. A $1 \times 1$ convolution layer first outputs an $H \times W$ binary foreground mask. Then, the standard FCN training pipeline is applied to train the network. Few (10) images per category were manually provided with groundtruth pixel-level foreground masks during training. During prediction, high-resolution foreground masks predicted by the network is fed to a CAM~\cite{camcvpr2016} procedure, which generates bounding boxes as localization results. \cite{evaluatecvpr2020} shows that even with only few groundtruth foreground mask supervision, this few-shot supervised method can beat state-of-the-art WSOL methods.

For WSOD, the method WSOD$^2$~\cite{wsod2iccv2019} uses external low-level objectness scores (including edge density and superpixel straddling) to inject extra objectness information into current WSOD methods. It calculates the low-level objectness score for each proposal, and filters away proposals with low scores during training. Performance gain by WSOD$^2$ shows that pixel-level objectness (foreground) scores, even obtained from low-level pixel information, can significantly boost the performance of WSOD methods. 

Thus, pixel-level groundtruth foreground masks must lead to even higher improvements for both WSOL and WSOD. This conjecture is verified by Figures~\ref{fig:wsol banner} and~\ref{fig:wsod banner}. For example, using the OICR method~\cite{oicrcvpr2017} with the VGG16 backbone, this HR GT information brings 3.3 mAP gains (from 46.9 of ``Weakly'' to 50.2 of ``HR GT'' in Fig.~\ref{fig:wsod banner}).

Pixel-level groundtruth information (even few-shot), however, are not available in WSOL. Low-level objectness scores for WSOD may contain high percentage of errors because it disregards image-level labels and high-level semantics. Hence, we propose weakly supervised foreground learning (WSFL), which \emph{learns foreground masks utilizing high-level deep learning techniques and image-level labels} in a weakly supervised fashion, and can be utilized in \emph{both WSOL and WSOD} as its downstream applications.

We propose a novel WSFL framework to learn how to estimate the unavailable groundtruth foreground masks, as shown in Algorithm~\ref{alg:WSFL}. In this algorithm, we do not need any groundtruth annotations, including bounding box annotations or pixel-level masks. 

First, we generate pseudo low resolution foreground masks. Then, using these (possibly highly noisy) masks as supervision signal, we train a low resolution foreground classification model, which predicts more accurate foreground masks for testing images. We will introduce these components step by step in the coming sections.

\begin{algorithm}[t]
	\caption{Weakly Supervised Foreground Learning}
	\begin{algorithmic} [1]
		\Statex {\textbf{Input}: Training images $I_{tr}$, testing images $I_{te}$}
		\Statex {\textbf{Output}: Foreground masks in testing images $P_{te}$}
		\State \quad Generate pseudo bounding boxes $\tilde{b}_{tr}$ on $I_{tr}$ 
		\State \quad Generate low resolution foreground masks $P_{tr}$ on $I_{tr}$ with $\tilde{b}_{tr}$ 
		\State \quad Train a low resolution foreground classification CNN $W_{pix}$ on $I_{tr}$ using $P_{tr}$ as supervision
		\State \quad Use $W_{pix}$ to predict $P_{te}$ for $I_{te}$
		\State \quad Upsample $P_{te}$ to the original size of the input images
		\State \textbf{Return:} $P_{te}$
	\end{algorithmic}
	\label{alg:WSFL}
\end{algorithm}

\subsection{Pseudo Foreground Mask Generation}

We now introduce how to generate pseudo foreground masks using only image-level labels. 

\subsubsection{High Resolution Pseudo Foreground Masks}

When groundtruth pixel annotations (\eg., for fully-supervised semantic segmentation) or bounding box annotations (\eg., for fully-supervised object detection) are available, it is straightforward to obtain high-resolution groundtruth foreground masks: A pixel is a foreground pixel (\ie, its mask value bing 1) if this pixel is inside any one of the object regions or bounding boxes; otherwise, the mask value is 0 (\ie, is a background pixel).

Since we are in a weakly supervised setting, we replace groundtruth bounding boxes with DDT~\cite{ddtpr2019} generated boxes. According to previous literature~\cite{psolcvpr2020}, we find that DDT~\cite{ddtpr2019}, an object co-localization method that only requires image-level labels, provides good pseudo bounding boxes that roughly separate objects from the background. We then generate high resolution pseudo foreground masks by setting a pixel’s foreground mask value to 1 if and only if it is within any one of these pseudo bounding boxes.

\subsubsection{Convert Pseudo Masks to Low Resolution}

Groundtruth high resolution masks has shown excellent utility for WSOL in~\cite{evaluatecvpr2020}. Similarly, high resolution masks in object detection is equivalent to fully-supervised detection, which outperforms WSOD significantly. But, the FCN-style network in~\cite{evaluatecvpr2020} requires high computational costs, which prevents it from being applied to large-scale datasets. DDT (and other co-localization methods) only generates one bounding box per image, which means our pseudo high resolution masks cannot be directly applied to WSOD.

To solve these difficulties, we turn the high resolution pseudo masks into low resolution ones, whose size $h \times w$ equals the spatial size of  the final activation maps in modern CNN models, e.g., $14 \times 14$. Then, we resize the bounding box annotations from inside the original input images to fit the low resolution ones. In turn, these bounding boxes are transformed into low resolution foreground masks. Low resolution maps will inevitably contain errors due to quantization issues. However, compared to high resolution semantic segmentation learning, its computational costs are significantly lower. Hence, it can be applied to large-scale datasets like the full training set of ImageNet-1k~\cite{imagenetijcv2015}.

\subsection{Low Resolution Pixel Classification}

An FCN-style semantic segmentation task using the low resolution pseudo masks as training labels are still time-consuming and difficult to optimize. Hence, we propose to use these pseudo masks as \emph{binary classification labels}, and learn a low resolution pixel classification model instead.

In detail, we simply take a widely used image classification model (such as VGG16~\cite{vggiclr2014} or ResNet50~\cite{resnetcvpr2016}), replace the final global averaging pooling and/or fully connected layers with a single $1\times1$ convolutional layer with only one output channel, leading to an output activation map of size $h\times w \times 1$ (same as that of the low resolution pseudo mask). We add a sigmoid layer to convert the output values to $[0,1]$, and minimize the binary cross entropy loss between these output values and the low resolution pseudo masks. Given a test image, the learned classifier predicts a binary map of size $h \times w$. Note that the predicted foreground map can have \emph{more than one disjoint foreground connected regions}, thus suitable for application in WSOD.

This mask can be easily resized to match the input image’s size using bilinear interpolation, which is \emph{what WSFL uses to replace the pixel-level groundtruth mask in a weakly supervised fashion}. 

Combining our low resolution mask and pixel classification model, we perform an oracle study in the WSOL task on ImageNet-1k with groundtruth foreground masks. Note that bounding box annotations on ImageNet-1k is incomplete, and we only use those images with groundtruth bounding box annotations to train the models. Experiment details can be found in Sec.~4. Results in Table~\ref{table:gt results on wsol} show that our ``low resolution mask + pixel classification model’’ outperforms the ``few-shot FCN-style network + high resolution mask’’ in~\cite{evaluatecvpr2020}. These result show the effectiveness of our proposed pipeline: converting high resolution masks to low resolution \emph{for all images} (followed by a pixel classification network) is not only efficient, but also more effective than using high resolution masks \emph{on only a few images}.

\begin{table}
	\small
	\centering
			\caption{Correct localization (CorLoc) accuracy of models in the WSOL task on ImageNet-1k. HR means high resolution foreground mask and LR means low resolution mask. }
			\vspace{-4pt}
			\label{table:gt results on wsol}
			\begin{tabular}{|c|r|c|}
				\hline
				Model  & Supervision & CorLoc\\
				\hline 
				ResNet50 & HR GT, few-shot~\cite{evaluatecvpr2020} & 67.5 \\
				ResNet50 & LR GT & 72.8 \\
				\hline
			\end{tabular}
\end{table}

\subsection{Downstream Applications of WSFL}

The foreground masks predicted by WSFL can be directly applied in WSOL tasks, since we can use CAM~\cite{camcvpr2016} to generate bounding box predictions. The utility of WSFL, however, is much wider than the simple WSOL.

In Sec~3.2, we show that groundtruth masks are useful for both WSOL/WSOD tasks. In Algorithm~\ref{alg:WSFL}, we train and predict foreground masks on the same dataset. However, since our WSFL model does \emph{not} need extra label inputs (unlike traditional CAM), we can directly transfer our model trained on a large-scale dataset (like ImageNet~\cite{imagenetijcv2015}) to different datasets (\eg, CUB-200~\cite{cubtech2011}) \emph{without modifications}.

Another natural application is WSOD. It is very difficult to find precise pseudo bounding boxes for multi-object datasets like PASCAL VOC~\cite{vocijcv2010}, when only image-level labels are available. Although there are weakly supervised semantic segmentation methods which can generate class-aware masks, their performances are still low and cannot meet our demand. Since we have shown that various groundtruth foreground (class-agnostic) masks can improve the performance of WSOD methods, and our WSFL model can transfer without any fine-tuning or additional labels, we can directly predict foreground masks in WSOD tasks using a \emph{pretrained} WSFL model.

After generating low resolution masks, we upsample them to the original $H \times W$ size using bilinear interpolation, and use the following process to calculate the foreground score for every proposal in the image. We treat the average value in the foreground mask of all pixels inside one object proposal as the objectness score for this proposal. Then, we add these scores into the WSOD training pipeline as extra inputs. State-of-the-art WSOD models will classify some input proposals as foreground proposals and use these foreground proposals to retrain the online classifier. In this paper, we follow the pipeline of WSOD$^2$: We classify some proposals as background proposals at the online instance refinement stage if they have objectness scores which are lower than a threshold. The detailed threshold value setting will be discussed in Sec~4.

One final note: Since all components in our WSFL (generating pseudo boxes using DDT, converting pseudo masks to low resolution, and pixel classification) have no interaction with the test set at all, the reasons described in~\cite{evaluatecvpr2020} for leading to an ill-posed problem do \emph{not} apply to WSFL.

\section{Experimental Setups and Details}

Next, we present our experimental setup for evaluating our WSFL framework, including the datasets and the implementation details.

\subsection{Datasets}

We evaluate our WSFL framework on two weakly supervised tasks: WSOL and WSOD. For WSOL, we use two standard datasets: ImageNet-1k~\cite{imagenetijcv2015} and CUB-200~\cite{cubtech2011}. ImageNet-1k is a single-object image classification dataset with around 1.3 million images. Bounding box annotations are incomplete for training images and complete for validation images. CUB-200 contains 200 kinds of different birds, with 11,788 images that have groundtruth box annotations. \emph{Except those lines with the ``GT'' suffix, we do not use any groundtruth annotation to train our WSFL model.} We use two metrics to evaluate our models: Top-1 localization accuracy (Top-1 Loc) and correct localization accuracy~(CorLoc). CorLoc is correct when given the class label to the WSOL model, the intersection over union~(IoU) score between the groundtruth box and the output box is larger than 50\% or more. Top-1 Loc is correct when the Top-1 classification and GT-Known Loc are both correct.

For WSOD, we will evaluate on a standard detection benchmark dataset VOC2007~\cite{vocijcv2010}. VOC2007 is an object detection dataset with 2,501 training images, 2,511 validation images and 4,952 testing images. We use training and validation images to train our models and test their performance on the testing images, following previous WSOD protocols~\cite{wsddncvpr2016,wsodsizeeccv2016,oicrcvpr2017}. During evaluation, we use the common mean average precision~(mAP) metric on test images.

\subsection{Implementation Details}

We use the PyTorch framework with 2080Ti GPUs to conduct our experiments. 

\textbf{Base Models.} For backbone models in WSOL, we use the same models used in previous methods~\cite{spgeccv2018,adlcvpr2019,evaluatecvpr2020}: VGG16~\cite{vggiclr2014}, InceptionV3~\cite{inceptionv3cvpr2016} and ResNet50~\cite{resnetcvpr2016}. For VGG16, we follow the previous guide in WSOL~\cite{camcvpr2016,spgeccv2018,adlcvpr2019}: remove the last max pooling layer to enlarge the receptive field. Also, we remove all fully connected layers in VGG16. For ResNet50, we remove the downsample stride of the last residual block. For InceptionV3, we follow the structural changes in~\cite{spgeccv2018,adlcvpr2019}. We use pretrained weights on ImageNet-1k to initialize our WSFL models. 

For backbone models in WSOD, we simply take the previous methods OICR and MIST~\cite{oicrcvpr2017,wetectroncvpr2020} as baseline models, then rerun these models with the extra supervision of our objectness scores.

For all experiments on WSOL, we first use DDT~\cite{ddtpr2019} to generate pseudo boxes on ImageNet-1k and CUB-200. Then, we generate pseudo foreground masks according to the pseudo boxes. We use the following hyperparameters on ImageNet: batch size 256, weight decay 0.0001. We use SGD optimizer with 0.9 momentum. For the learning rate, we will start at 0.001, then decay at every 4 epochs. The total training epochs on ImageNet is 12 for all models. For WSOL experiments on CUB-200, we keep other hyperparameters the same and change the batch size from 256 to 64. The learning rate will decay at every 10 epochs and the total training epochs is 30 on CUB-200. We directly resize all input images into $224\times224$, then perform random horizontal flipping for training. For testing, we directly resize the input image to $224\times224$, then feed into the model.

After getting the low resolution output, we will use bilinear interpolation to upsample the low resolution output to the size of the original input image. Then we will use CAM~\cite{camcvpr2016} to generate the output bounding box. Since our models can directly output localization results, we follow the instructions in~\cite{psolcvpr2020} to combine localization results with classification results. For results on CUB-200, we find that fine-tuning WSFL models trained on ImageNet-1k will have better performance. Ablation studies on different initialization weights on CUB-200 will be presented in Sec.~5.

For WSOD experiments, we directly use a ResNet50 WSFL model pre-trained on the ImageNet dataset to generate pseudo foreground masks on VOC2007 without any fine-tuning. We use the same hyperparameters and evaluation pipeline of OICR and MIST, and do not make any further changes. For proposal filtering, we set threshold as 0.2 to filter proposals with our WSFL masks. For groundtruth masks, we set the threshold as 0.5 when we apply our WSFL. Visual inspections find that our WSFL model performs poorly on the ``person’’ and ``plant’’ categories on VOC2007, possibly because these categories do not appear in ImageNet. Thus, we do not filter proposals which are classified as foreground of person and plant in the online instance refinement stage.

\section{Results and Analyses}

In this section, we provide empirical results and analyses of the proposed WSFL model and applications. 

\subsection{WSOL Results}

First, we show results on WSOL benchmark datasets and compare with state-of-the-art WSOL methods. Top-1 Loc and CorLoc results are shown in Table~\ref{table:all clsloc results}. From the table, we have the following observations and findings.

\begin{table}[t]
	\scriptsize
	\centering
			\caption{Top-1 localization and CorLoc results on CUB-200 and ImageNet-1k. VGG-GAP means we replace three fully connected layers in VGG16 with global average pooling and one fully connected layer, and for VGG16 it keeps the original VGG16 structure. For compared methods, we report numbers in previous papers. ``-'' means results were not reported in the respective paper. ``LR GT'' means the model are trained with low resolution groundtruth foreground masks. Best results are shown in boldface. An important note: For few-shot GT FCN~\cite{evaluatecvpr2020} and LR GT, since they rely on extra supervision signals, their results are listed here to provide a context, and they are not counted in the comparison.}
			\label{table:all clsloc results}
			\setlength{\tabcolsep}{2.5pt}
			\vspace{-4pt}
			\begin{tabular}{|l|r|r|r|r|r|r|r|}
				\hline
				\multirow{2}{*}{Model}  & \multirow{2}{*}{Backbone} &\multicolumn{2}{|c|}{CUB-200}&\multicolumn{2}{|c|}{ImageNet-1k}\\
				\cline{3-4} \cline{5-6}
				& & Top-1 Loc  & CorLoc & Top-1 Loc & CorLoc\\
				\hline 
				
				VGG16-CAM~\cite{camcvpr2016} & VGG-GAP & 37.05 & 53.68 & 42.80 & 59.00\\
				VGG16-ACoL~\cite{acolcvpr2018} & VGG-GAP  & 45.92 & - & 45.83 & 62.96\\
				ADL~\cite{adlcvpr2019} & VGG-GAP & 53.40 & 73.96 & 42.96 & 59.24 \\
				CutMix~\cite{cutmixiccv2019} & VGG-GAP & 52.53 & - & 43.45 & - \\
				DDT~\cite{ddtpr2019} & VGG16 & 62.30 & 84.55 & 47.31 & 61.41\\ 
				PSOL-Sep~\cite{psolcvpr2020} & VGG-GAP & 59.29 & 80.45 & 48.36 & 63.72\\
				I$^2$C~\cite{i2ceccv2020} & VGG-GAP & 55.99 & 72.60  & 47.41 & 63.90 \\
				SEM~\cite{semarxiv2020} & VGG-GAP & - & -  & 47.53 & 63.47 \\
				CAM w/~\cite{rethinkcameccv2020} & VGG-GAP & 61.30 & 80.72 & 45.40 & 62.68 \\
				WSFL & VGG-GAP &\textbf{68.33} & \textbf{92.92}  & \textbf{51.47} & \textbf{66.95}\\
				\hline
				Few-Shot GT FCN~\cite{evaluatecvpr2020} & VGG16 & - & 86.30 & - & 62.80\\
				
				\hline
				\hline 
				SPG~\cite{spgeccv2018} & InceptionV3  &  46.64 & - & 48.60  & 64.69\\
				ADL~\cite{adlcvpr2019} & InceptionV3 & 53.04 & - & 48.71 & - \\
				ADL w/~\cite{rethinkcameccv2020} & InceptionV3 & 53.04 & 69.95 & 50.56 & 64.44 \\
				PSOL-Sep~\cite{psolcvpr2020} & InceptionV3 & 65.51 & 83.44 & 54.82  & 65.21 \\
				I$^2$C~\cite{i2ceccv2020} & InceptionV3 & 55.99 & 72.60  &53.17 & 68.50 \\
				SEM~\cite{semarxiv2020} & InceptionV3 & - & -  & 53.04 & 69.04 \\
				WSFL & InceptionV3 & \textbf{69.04} & \textbf{93.96} & \textbf{57.12} & \textbf{69.59}\\
				\hline
				Few-Shot GT FCN~\cite{evaluatecvpr2020} & InceptionV3 & - & {94.00} & - & 68.70\\

				\hline \hline 
				ADL~\cite{adlcvpr2019} & ResNet50-SE & 62.29 & - & 48.53  & -\\
				ADL w/~\cite{rethinkcameccv2020} & ResNet50 & 59.53 & 77.58 & 49.42 & 62.20 \\
				CutMix~\cite{cutmixiccv2019} & ResNet50 & 54.81 & - & 47.25 & - \\
				I$^2$C~\cite{i2ceccv2020} & ResNet50 & - & -  &54.83 & 68.50 \\
				PSOL~\cite{psolcvpr2020} & ResNet50  & 69.87 & 86.56 & 53.98 & 65.44\\
	                WSFL & ResNet50 & \textbf{72.97} & \textbf{94.75} & \textbf{56.56} & \textbf{69.53}\\ 
	                \hline 
	                Few-Shot GT FCN~\cite{evaluatecvpr2020} & ResNet50 & - & {95.80} & - & 67.50\\
	                LR GT & ResNet50 & 75.87& 98.45 & 59.30& 72.80\\
	                \hline 
			\end{tabular}

\end{table}

\squishlist
	\item WSFL performs better than baseline methods with the same inputs, including DDT~\cite{ddtpr2019} and PSOL~\cite{psolcvpr2020}. From the performance on VGG-GAP backbone, we can see that WSFL performs significantly better than DDT and PSOL on both datasets. Compared to PSOL~\cite{psolcvpr2020}, WSFL turns the pseudo box annotations into low resolution foreground masks and obtains better results. This phenomenon shows we should utilize more information than bounding box coordinates in WSOL tasks.
	
	\item WSFL significantly outperforms other WSOL methods with the same level of supervision. With the same post-processing step (CAM) and same backbone~\cite{camcvpr2016}, WSFL outperforms previous WSOL methods~\cite{adlcvpr2019,spgeccv2018,cutmixiccv2019,i2ceccv2020} by a large gap. With the same pseudo supervision, we have better performance than DDT~\cite{ddtpr2019} and PSOL~\cite{psolcvpr2020}. Some recent methods try to modify CAM~\cite{camcvpr2016} to achieve better performance~\cite{semarxiv2020,rethinkcameccv2020}. Our WSFL models, even without any modification to CAM, still achieve significantly better accuracy than these methods. Moreover, WSFL can be combined with these post-processing methods to further achieve better performance.
	
	\item WSFL can achieve comparable or better performance than few-shot baselines, which breaks the claim in~\cite{evaluatecvpr2020}. Our WSFL models have comparable or better performance than few-shot FCNs. In~\cite{evaluatecvpr2020}, they mentioned that WSOL is an ill-posed problem, and researchers should use few-shot learning to achieve better performance. However, our WSFL framework shows that, so long as we utilize proper inputs in the current WSOL setting, we can achieve comparable or better performance than few-shot FCN. Consider the scalability issues of FCN-style networks on large-scale datasets like ImageNet-1k, WSFL with its low resolution classification model can predict and scale better.
\squishend

\subsection{WSOD Results}

Now we move on to introduce results of WSFL on WSOD, summarized in Table~\ref{table:wsfl results on wsod}. Our results suggest that:
\begin{table}
	\small
	\centering
			\caption{mAP results of different WSOD methods on VOC2007 with extra information given by WSFL and/or other models. Please note that we re-implemented the OICR~\cite{oicrcvpr2017} and MIST~\cite{wetectroncvpr2020} baseline methods in their respective paper, and our results are higher than results reported in the original papers. The column ``WS’’ denotes whether a method is weakly supervised or not. Both ``HR GT’’ and ``LR GT’’ use groundtruth pixel supervision, and should \emph{not} be directly compared with WSFL.}
			\label{table:wsfl results on wsod}
			\vspace{-6pt}
			\setlength{\tabcolsep}{2pt}
			\begin{tabular}{|c|c|r|c|}
				\hline
				Baseline method  & WS  & Extra information  & mAP\\
				\hline
				WS-JDS & \cmark & WS-JDS~\cite{wsjdscvpr2019} & 45.6 \\
				\hline 
				\multirow{5}{*}{OICR} & \cmark & None  & 46.9\\
			        & \cmark & SDCN~\cite{sdcniccv2019} & 46.0 \\
			        & \cmark & OCRepr~\cite{ocreprmm2020} & 46.3 \\
				 	& \cmark & WSOD$^2$~\cite{wsod2iccv2019} & 48.1 \\
				 	& \cmark & WSFL & \textbf{48.3}\\ \cline{2-4}
					& \xmark & LR GT & 48.6 \\
					& \xmark  & HR GT & 50.2 \\
				 \hline \hline
				\multirow{4}{*}{MIST} & \cmark & None & 55.2 \\
					 & \cmark & WSFL & \textbf{55.7} \\ \cline{2-4}
					 & \xmark & LR GT & 56.5 \\
					 & \xmark & HR GT & 56.8 \\
				\hline
			\end{tabular}
	\vspace{-6pt}
\end{table}

\squishlist
    \item With groundtruth foreground pixel masks, various WSOD methods get consistent improvements over their baselines. For two recent WSOD methods, OICR~\cite{oicrcvpr2017} and MIST~\cite{wetectroncvpr2020}, high resolution groundtruth mask scores can improve 3.3 mAP and 1.4 mAP for them, respectively. With low resolution groundtruth masks, we can still have 1.7 mAP and 1.2 mAP gains, respectively, which proves that even low resolution masks can still boost the performance of various WSOD methods.
    \item Our WSFL model can provide a substitution for groundtruth foreground masks. Masks predicted by WSFL lead to 1.3 mAP and 0.5 mAP gain on these WSOD methods, respectively. which is lower than, but close to gains by low resolution groundtruth masks. 
    \item Our WSFL achieves higher mAP than WSOD methods with extra segmentation branches~\cite{sdcniccv2019,ocreprmm2020,wsjdscvpr2019} and slightly higher mAP than the method using low-level vision scores~\cite{wsod2iccv2019}. In the mean time, computing the foreground mask in WSFL is more efficient than computing the segmentation mask. In the future, the synergy between WSFL masks and other scores may further improve WSOD.
\squishend

\subsection{Good Transfer Ability}

In this section, we will provide analyses on the transfer ability of our WSFL models. 

Previously, WSOL methods require class labels to generate bounding boxes, which cannot be transferred between different datasets. PSOL~\cite{psolcvpr2020} shows that their bounding box prediction models trained on ImageNet can achieve good performance without any further fine-tuning on a different dataset. We want to verify whether our WSFL model has good transfer ability, too. In fact, since we used a ImageNet pretrained WSFL model in WSOD tasks (on Pascal VOC), the WSFL transfer ability has been indirectly validated.

\begin{table}
	\small
	\centering
			\caption{Correct localization~(CorLoc) accuracy of different WSFL models with different initial weights and training targets on CUB-200.}
			\label{table:transfer results on wsol}
			\vspace{-6pt}
			\setlength{\tabcolsep}{2pt}
			\begin{tabular}{|c|r|c|c|}
				\hline
				Model  & Initialization Weights & Fine-tuned & CorLoc\\
				\hline 
				\multirow{3}{*}{ResNet50} & WSFL on ImageNet & \xmark & 91.25\\
					 & Classification & \cmark & 93.55 \\
					 & WSFL on ImageNet & \cmark & 94.75 \\
				\hline 
				\multirow{3}{*}{VGG-GAP} & WSFL on ImageNet & \xmark & 82.59\\
					 & Classification & \cmark & 91.18\\
					 & WSFL on ImageNet & \cmark & 92.92 \\
				\hline
			     \multirow{3}{*}{InceptionV3} & WSFL on ImageNet & \xmark & 90.92\\
					 & Classification & \cmark & 91.25 \\
					 & WSFL on ImageNet & \cmark & 93.96 \\
				\hline
			\end{tabular}
	\vspace{-6pt}
\end{table}

\begin{figure*}
	\centering
	\includegraphics[width=0.78\linewidth]{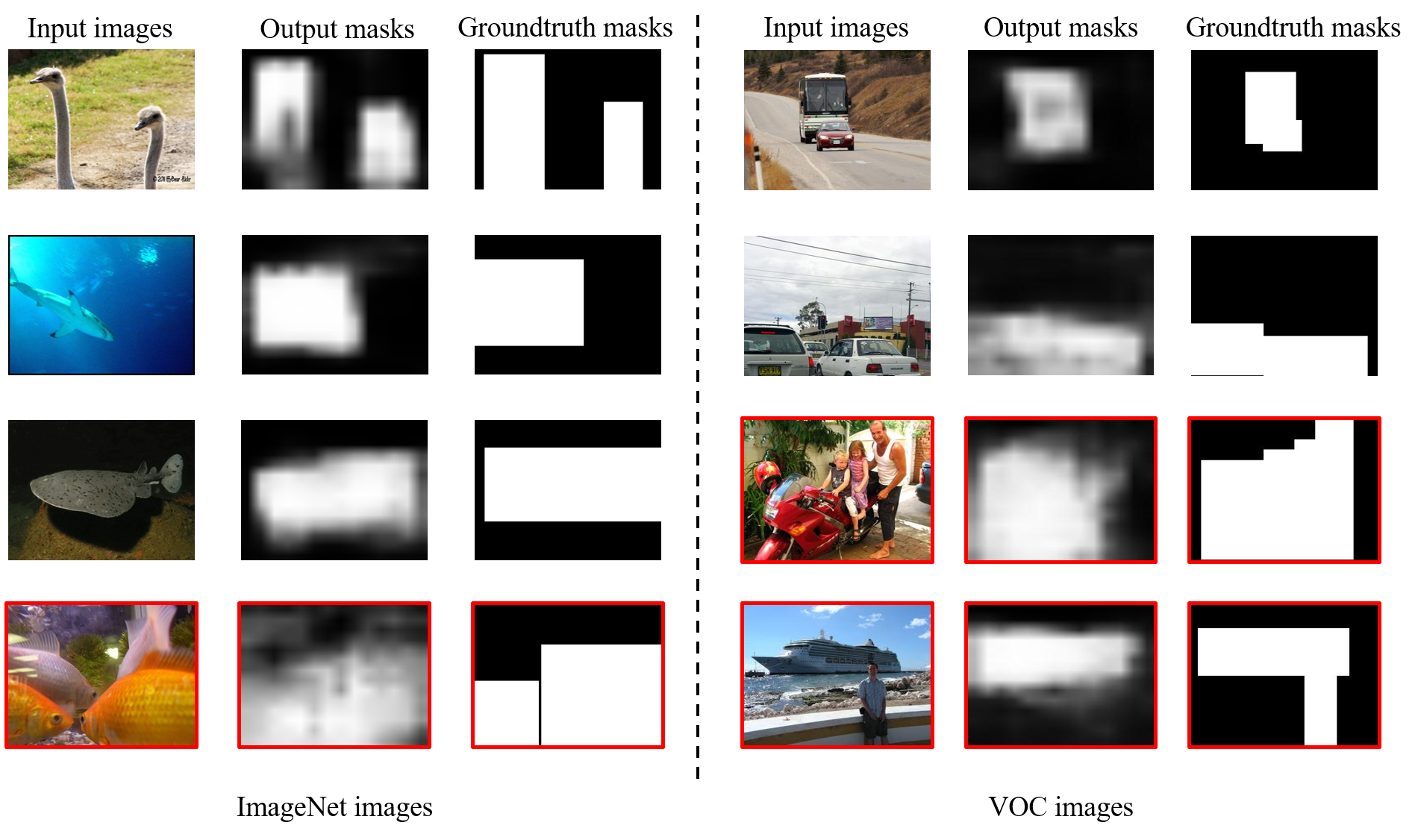}
	\vspace{-8pt}
	\caption{Visualization of our WSFL model's output on images from different datasets. On the left, we randomly choose 4 images from the ImageNet-1k validation dataset; on the right, since we do not perform any \textit{fine-tuning}, we randomly choose 4 images from the VOC 2007 trainval dataset. For every image, we show the foreground masks predicted by our WSFL model, along with the groundtruth pixel-level foreground mask provided (or transferred) from annotations. The ResNet50 WSFL model was trained on ImageNet. Pictures with red edges mean that our WSFL models output incorrect masks for the input image. For example, as aforementioned, ``person’’ in VOC images are often missing in the predicted masks.}
	\label{fig:visualization results}
\end{figure*}

We take all baseline models in Table~\ref{table:all clsloc results} and use CUB-200 as the target dataset to explore WSFL’s transfer ability between different datasets. There are two weight initialization choices for WSFL models: Classification weights on ImageNet, and WSFL weights on ImageNet. Furthermore, for WSFL weights on ImageNet, we can choose whether to fine-tune it on the pseudo CUB foreground masks generated by DDT. The results are in Table~\ref{table:transfer results on wsol}. From the table we have the following conclusions.

\squishlist
	\item Without any fine-tuning, WSFL models perform well on CUB-200, except in the VGG-GAP model which have a large gap~(10\%) with fine-tuned models. ResNet50 and InceptionV3 only have small gaps, i.e., about 3\% CorLoc accuracy. These phenomena show that our WSFL models transfer well on different WSOL datasets. That also explains why we can apply our WSFL models on WSOD methods to boost its accuracy.
	\item WSFL models provide better initialization compared with classification weights. Compared to classification weights, all three models show 1-2\% performance gain when WSFL weights are fine-tuned. These results show that there are indeed some characteristic differences between the classification and the localization task, which has been shown in~\cite{psolcvpr2020}, too. We need to use separate models to conduct classification and localization tasks. 
\squishend

\subsection{Visualization and Failure Cases}

In this section, we show some visualization results to evaluate the output of our WSFL models. The visualizations are in Fig.~\ref{fig:visualization results}. Moreover, we provide the output of our WSFL models and weakly supervised saliency detection methods in the supplementary material.

Fig.~\ref{fig:visualization results} clearly show that WSFL models output precise masks for object localization tasks, even with multiple separate objects (the bottom-left image). The masks are also high-quality when transferred to a different dataset without fine-tuning (top 2 images in the right half). 

However, our WSFL models fail to predict the bounding boxes for nearby objects, although we have correct foreground masks. The problem comes from the CAM post-processing part. Also, since ImageNet-1k does not have the ``person'' category, people are often labeled as background in WSFL ImageNet learning. Thus, the learned WSFL model will not label person in the image as foreground. We need to deal with these biases in the future.

\section{Conclusion and Future Work}

In this paper, we established the motivation and necessity for weakly supervised foreground learning (WSFL), and proposed the WSFL task. We also showed that a successful WSFL model is very valuable for various downstream applications, such as weakly supervised object localization and detection (WSOL and WSOD). We then proposed a computationally efficient WSFL pipeline. Our WSFL model significantly improves the performance of WSOL and WSOD, and has demonstrated excellent transfer ability. We believe that our work provide a solid step towards weakly supervised tasks in computer vision.

In the future, we will explore better backbones for WSFL (\eg, removing biases), and will explore more applications of WSFL in different tasks, including better proposal scoring for WSOD tasks, and application in other new tasks (such as weakly supervised semantic segmentation).

\clearpage
{\small
\bibliographystyle{ieee_fullname}
\bibliography{egbib}

\begin{thebibliography}{10}\itemsep=-1pt

\bibitem{rethinkcameccv2020}
Wonho Bae, Junhyug Noh, and Gunhee Kim.
\newblock Rethinking class activation mapping for weakly supervised object
  localization.
\newblock In {\em Eur. Conf. Comput. Vis.}, volume in press of {\em LNCS}, page
  in press, 2020.

\bibitem{wsddncvpr2016}
Hakan Bilen and Andrea Vedaldi.
\newblock Weakly supervised deep detection networks.
\newblock In {\em IEEE Conf. Comput. Vis. Pattern Recog.}, pages 2846--2854,
  2016.

\bibitem{saliencylataaai2018}
Chunshui Cao, Yongzhen Huang, Zilei Wang, Liang Wang, Ninglong Xu, and Tieniu
  Tan.
\newblock Lateral inhibition-inspired convolutional neural network for visual
  attention and saliency detection.
\newblock In {\em AAAI}, pages 6690--6697, 2018.

\bibitem{evaluatecvpr2020}
Junsuk Choe, Seong~Joon Oh, Seungho Lee, Sanghyuk Chun, Zeynep Akata, and
  Hyunjung Shim.
\newblock Evaluating weakly supervised object localization methods right.
\newblock In {\em IEEE Conf. Comput. Vis. Pattern Recog.}, pages 3133--3142,
  2020.

\bibitem{adlcvpr2019}
Junsuk Choe and Hyunjung Shim.
\newblock Attention-based dropout layer for weakly supervised object
  localization.
\newblock In {\em IEEE Conf. Comput. Vis. Pattern Recog.}, pages 2219--2228,
  2019.

\bibitem{vocijcv2010}
Mark Everingham, Luc Van~Gool, Christopher~KI Williams, John Winn, and Andrew
  Zisserman.
\newblock The {PASCAL} visual object classes ({VOC}) challenge.
\newblock {\em Int. J. Comput. Vis.}, 88(2):303--338, 2010.

\bibitem{resnetcvpr2016}
Kaiming He, Xiangyu Zhang, Shaoqing Ren, and Jian Sun.
\newblock Deep residual learning for image recognition.
\newblock In {\em IEEE Conf. Comput. Vis. Pattern Recog.}, pages 770--778,
  2016.

\bibitem{contextlocneteccv2016}
Vadim Kantorov, Maxime Oquab, Minsu Cho, and Ivan Laptev.
\newblock {ContextLocNet}: Context-aware deep network models for weakly
  supervised localization.
\newblock In {\em Eur. Conf. Comput. Vis.}, volume 9909 of {\em LNCS}, pages
  350--365, 2016.

\bibitem{saliencyvaeaaai2019}
Bo Li, Zhengxing Sun, and Yuqi Guo.
\newblock {SuperVAE}: Superpixelwise variational autoencoder for salient object
  detection.
\newblock In {\em AAAI}, pages 8569--8576, 2019.

\bibitem{sdcniccv2019}
Xiaoyan Li, Meina Kan, Shiguang Shan, and Xilin Chen.
\newblock Weakly supervised object detection with segmentation collaboration.
\newblock In {\em Int. Conf. Comput. Vis.}, pages 9735--9744, 2019.

\bibitem{fcncvpr2015}
Jonathan Long, Evan Shelhamer, and Trevor Darrell.
\newblock Fully convolutional networks for semantic segmentation.
\newblock In {\em IEEE Conf. Comput. Vis. Pattern Recog.}, pages 3431--3440,
  2015.

\bibitem{eilcvpr2020}
Jinjie Mai, Meng Yang, and Wenfeng Luo.
\newblock Erasing integrated learning: A simple yet effective approach for
  weakly supervised object localization.
\newblock In {\em IEEE Conf. Comput. Vis. Pattern Recog.}, pages 8766--8775,
  2020.

\bibitem{wetectroncvpr2020}
Zhongzheng Ren, Zhiding Yu, Xiaodong Yang, Ming-Yu Liu, Yong~Jae Lee,
  Alexander~G Schwing, and Jan Kautz.
\newblock Instance-aware, context-focused, and memory-efficient weakly
  supervised object detection.
\newblock In {\em IEEE Conf. Comput. Vis. Pattern Recog.}, pages 10598--10607,
  2020.

\bibitem{imagenetijcv2015}
Olga Russakovsky, Jia Deng, Hao Su, Jonathan Krause, Sanjeev Satheesh, Sean Ma,
  Zhiheng Huang, Andrej Karpathy, Aditya Khosla, Michael Bernstein,
  Alexander~C. Berg, and Li Fei-Fei.
\newblock {ImageNet} large scale visual recognition challenge.
\newblock {\em Int. J. Comput. Vis.}, 115(3):211--252, 2015.

\bibitem{gradcamiccv2017}
Ramprasaath~R Selvaraju, Michael Cogswell, Abhishek Das, Ramakrishna Vedantam,
  Devi Parikh, and Dhruv Batra.
\newblock {Grad-CAM}: Visual explanations from deep networks via gradient-based
  localization.
\newblock In {\em Int. Conf. Comput. Vis.}, pages 618--626, 2017.

\bibitem{enableresneteccv2020}
Yunhang Shen, Rongrong Ji, Yan Wang, Zhiwei Chen, Feng Zheng, Feiyue Huang, and
  Yunsheng Wu.
\newblock Enabling deep residual networks for weakly supervised object
  detection.
\newblock In {\em Eur. Conf. Comput. Vis.}, volume 12353 of {\em LNCS}, pages
  118--136, 2020.

\bibitem{wsjdscvpr2019}
Yunhang Shen, Rongrong Ji, Yan Wang, Yongjian Wu, and Liujuan Cao.
\newblock Cyclic guidance for weakly supervised joint detection and
  segmentation.
\newblock In {\em IEEE Conf. Comput. Vis. Pattern Recog.}, pages 697--707,
  2019.

\bibitem{wsodsizeeccv2016}
Miaojing Shi and Vittorio Ferrari.
\newblock Weakly supervised object localization using size estimates.
\newblock In {\em Eur. Conf. Comput. Vis.}, volume 9909 of {\em LNCS}, pages
  105--121, 2016.

\bibitem{vggiclr2014}
Karen Simonyan and Andrew Zisserman.
\newblock Very deep convolutional networks for large-scale image recognition.
\newblock In {\em Int. Conf. Learn. Represent.}, pages 1--14, 2015.

\bibitem{hideandseekiccv2017}
Krishna~Kumar Singh and Yong~Jae Lee.
\newblock {Hide-and-Seek}: Forcing a network to be meticulous for
  weakly-supervised object and action localization.
\newblock In {\em Int. Conf. Comput. Vis.}, pages 3544--3553, 2017.

\bibitem{inceptionv3cvpr2016}
Christian Szegedy, Vincent Vanhoucke, Sergey Ioffe, Jon Shlens, and Zbigniew
  Wojna.
\newblock Rethinking the {Inception} architecture for computer vision.
\newblock In {\em IEEE Conf. Comput. Vis. Pattern Recog.}, pages 2818--2826,
  2016.

\bibitem{pcltpami2018}
Peng Tang, Xinggang Wang, Song Bai, Wei Shen, Xiang Bai, Wenyu Liu, and Alan
  Yuille.
\newblock {PCL}: Proposal cluster learning for weakly supervised object
  detection.
\newblock {\em IEEE Trans. Pattern Anal. Mach. Intell.}, 42(1):176--191, 2018.

\bibitem{oicrcvpr2017}
Peng Tang, Xinggang Wang, Xiang Bai, and Wenyu Liu.
\newblock Multiple instance detection network with online instance classifier
  refinement.
\newblock In {\em IEEE Conf. Comput. Vis. Pattern Recog.}, pages 2843--2851,
  2017.

\bibitem{cubtech2011}
C. Wah, S. Branson, P. Welinder, P. Perona, and S. Belongie.
\newblock {The Caltech-UCSD birds-200-2011 dataset}.
\newblock Technical Report CNS-TR-2011-001, California Institute of Technology,
  2011.

\bibitem{cmilcvpr2019}
Fang Wan, Chang Liu, Wei Ke, Xiangyang Ji, Jianbin Jiao, and Qixiang Ye.
\newblock {C-MIL}: Continuation multiple instance learning for weakly
  supervised object detection.
\newblock In {\em IEEE Conf. Comput. Vis. Pattern Recog.}, pages 2199--2208,
  2019.

\bibitem{wsscvpr2017}
Lijun Wang, Huchuan Lu, Yifan Wang, Mengyang Feng, Dong Wang, Baocai Yin, and
  Xiang Ruan.
\newblock Learning to detect salient objects with image-level supervision.
\newblock In {\em IEEE Conf. Comput. Vis. Pattern Recog.}, pages 136--145,
  2017.

\bibitem{saliencytpami2021}
Wenguan Wang, Qiuxia Lai, Huazhu Fu, Jianbing Shen, Haibin Ling, and Ruigang
  Yang.
\newblock Salient object detection in the deep learning era: an in-depth
  survey.
\newblock {\em IEEE Trans. Pattern Anal. Mach. Intell.}, in press:in press,
  2021.

\bibitem{ddtpr2019}
Xiu-Shen Wei, Chen-Lin Zhang, Jianxin Wu, Chunhua Shen, and Zhi-Hua Zhou.
\newblock Unsupervised object discovery and co-localization by deep descriptor
  transformation.
\newblock {\em Pattern Recognition}, 88:113--126, 2019.

\bibitem{saliencyref2iccv2019}
Zhe Wu, Li Su, and Qingming Huang.
\newblock Stacked cross refinement network for edge-aware salient object
  detection.
\newblock In {\em Int. Conf. Comput. Vis.}, pages 7264--7273, 2019.

\bibitem{ocreprmm2020}
Ke Yang, Peng Zhang, Peng Qiao, Zhiyuan Wang, Dongsheng Li, and Yong Dou.
\newblock Objectness consistent representation for weakly supervised object
  detection.
\newblock In {\em ACM Int. Conf. Multimedia}, pages 1688--1696, 2020.

\bibitem{cutmixiccv2019}
Sangdoo Yun, Dongyoon Han, Seong~Joon Oh, Sanghyuk Chun, Junsuk Choe, and
  Youngjoon Yoo.
\newblock {CutMix}: Regularization strategy to train strong classifiers with
  localizable features.
\newblock In {\em Int. Conf. Comput. Vis.}, pages 6023--6032, 2019.

\bibitem{saliencyref1iccv2019}
Yu Zeng, Yunzhi Zhuge, Huchuan Lu, and Lihe Zhang.
\newblock Joint learning of saliency detection and weakly supervised semantic
  segmentation.
\newblock In {\em Int. Conf. Comput. Vis.}, pages 7223--7233, 2019.

\bibitem{wsod2iccv2019}
Zhaoyang Zeng, Bei Liu, Jianlong Fu, Hongyang Chao, and Lei Zhang.
\newblock {WSOD$^2$}: Learning bottom-up and top-down objectness distillation
  for weakly-supervised object detection.
\newblock In {\em Int. Conf. Comput. Vis.}, pages 8292--8300, 2019.

\bibitem{psolcvpr2020}
Chen-Lin Zhang, Yun-Hao Cao, and Jianxin Wu.
\newblock Rethinking the route towards weakly supervised object localization.
\newblock In {\em IEEE Conf. Comput. Vis. Pattern Recog.}, pages 13460--13469,
  2020.

\bibitem{saliencyref3cvpr2019}
Lu Zhang, Jianming Zhang, Zhe Lin, Huchuan Lu, and You He.
\newblock {CapSal}: Leveraging captioning to boost semantics for salient object
  detection.
\newblock In {\em IEEE Conf. Comput. Vis. Pattern Recog.}, pages 6024--6033,
  2019.

\bibitem{acolcvpr2018}
Xiaolin Zhang, Yunchao Wei, Jiashi Feng, Yi Yang, and Thomas~S Huang.
\newblock Adversarial complementary learning for weakly supervised object
  localization.
\newblock In {\em IEEE Conf. Comput. Vis. Pattern Recog.}, pages 1325--1334,
  2018.

\bibitem{spgeccv2018}
Xiaolin Zhang, Yunchao Wei, Guoliang Kang, Yi Yang, and Thomas Huang.
\newblock Self-produced guidance for weakly-supervised object localization.
\newblock In {\em Eur. Conf. Comput. Vis.}, volume 11216 of {\em LNCS}, pages
  610--625, 2018.

\bibitem{i2ceccv2020}
Xiaolin Zhang, Yunchao Wei, and Yi Yang.
\newblock Inter-image communication for weakly supervised localization.
\newblock In {\em Eur. Conf. Comput. Vis.}, volume 12364 of {\em LNCS}, pages
  271--287, 2020.

\bibitem{semarxiv2020}
Xiaolin Zhang, Yunchao Wei, Yi Yang, and Fei Wu.
\newblock Rethinking localization map: Towards accurate object perception with
  self-enhancement maps.
\newblock {\em arXiv preprint arXiv:2006.05220}, 2020.

\bibitem{camcvpr2016}
Bolei Zhou, Aditya Khosla, Agata Lapedriza, Aude Oliva, and Antonio Torralba.
\newblock Learning deep features for discriminative localization.
\newblock In {\em IEEE Conf. Comput. Vis. Pattern Recog.}, pages 2921--2929,
  2016.

\end{thebibliography}
}

\clearpage

\section*{Supplementary Materials}
This is the supplementary material for our paper, titled ``Weakly Supervised Foreground Learning for Weakly Supervised Localization and Detection''. In this document, we provide additional visualization and comparison with weakly supervised saliency methods.

\subsection*{Comparison with Saliency Methods}
We now present the comparison between our method and weakly supervised saliency methods.

For weakly supervised saliency methods, we choose WSS~\cite{wsscvpr2017}, which trains a saliency detection model with image-level supervision. We list the difference between our WSFL model and WSS as follows: WSFL aims to learn low-resolution foreground masks with a classification backbone while WSS learns high-resolution saliency masks based on a segmentation backbone. Thus, WSFL can be trained on a large-scale dataset like the full ImageNet-1k~\cite{imagenetijcv2015}, which contains more than one million images. However, WSS can only be trained on a small dataset. In the WSS paper, they used a subset of ImageNet-DET, which only contains around 10,000 images.

For a detailed comparison, we use WSS to replace proposal scores generated by WSFL models. Then we use these proposals to conduct weakly supervised object detection tasks on VOC 2007. The results are in Table~\ref{table:wsfl results with wss}. From Table~\ref{table:wsfl results with wss}, we can see that, our WSFL model can achieve better result than WSS, which indicates the effectiveness of our WSFL model.

\begin{table}
	\small
	\centering
			\caption{mAP results of different WSOD methods on VOC2007 with extra information given by WSFL and WSS models. Please note that we re-implemented the OICR~\cite{oicrcvpr2017} baseline methods in their respective paper, and our results are higher than results reported in the original papers.}
			\label{table:wsfl results with wss}
			\setlength{\tabcolsep}{2pt}
			\begin{tabular}{|c|r|c|}
				\hline
				Baseline method    & Extra information  & mAP\\
				\hline
				\multirow{3}{*}{OICR} & None  & 46.9\\
			        & WSS~\cite{wsscvpr2017} & 47.5 \\
				 	& WSFL & \textbf{48.3}\\ \hline

			\end{tabular}
	\vspace{-6pt}
\end{table}

Moreover, we present the visualization of WSFL and WSS on the PASCAL VOC2007~\cite{vocijcv2010} dataset. For WSS, we directly take pre-trained models from the official website.\footnote{https://github.com/scott89/WSS} For WSFL, we directly take a ResNet50 WSFL model pre-trained on the ImageNet dataset.
\begin{figure*}
	\centering
	\includegraphics[width=0.9\linewidth]{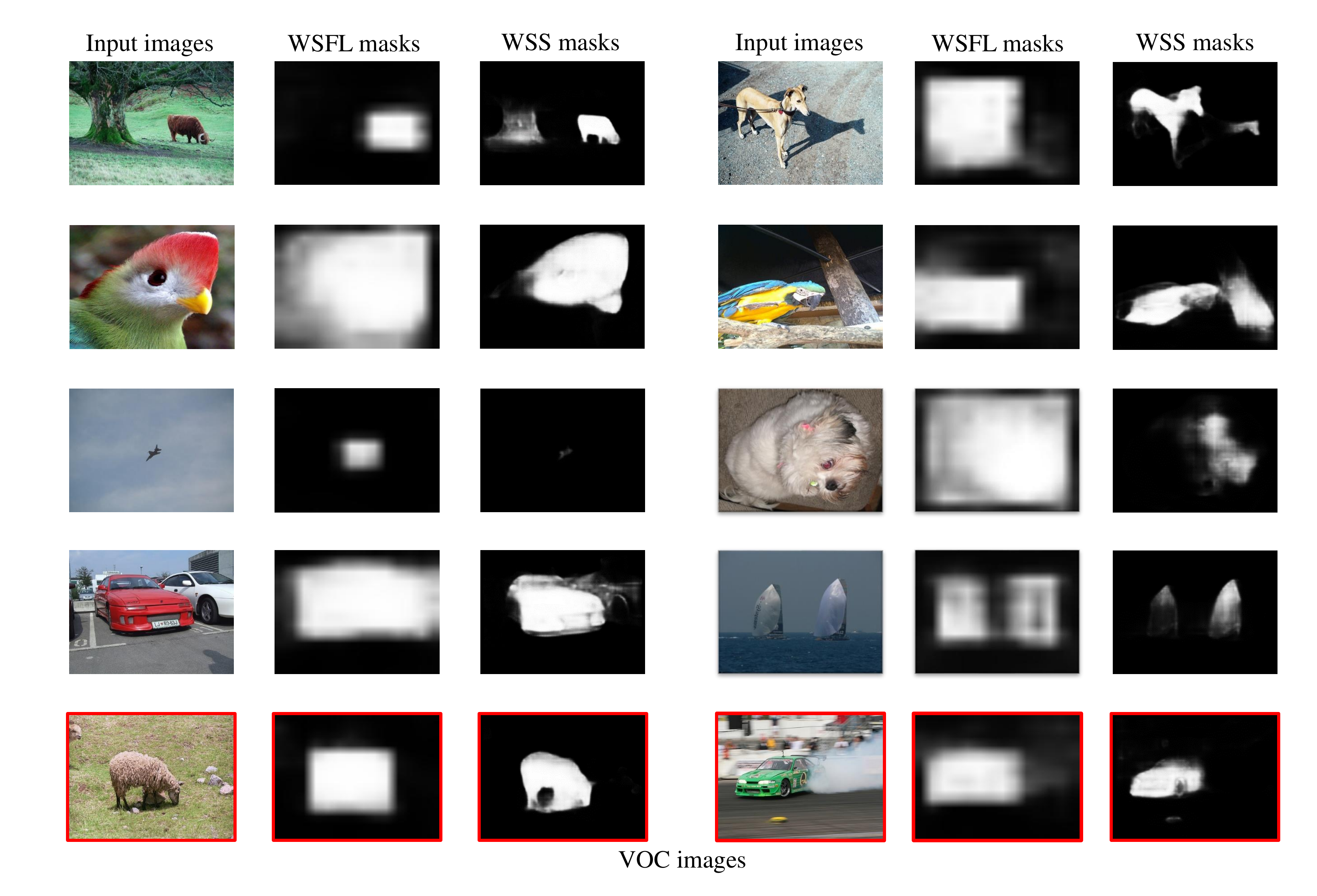}
	\caption{Visualization of our WSFL model's output and WSS's output on randomly chosen images from the VOC2007 dataset. Please note that, WSFL and WSS are not trained or finetuned on the VOC2007 dataset. We directly evaluate these models. The first to third columns show the outputs of the first 5 images and the fourth to sixth columns show the outputs of another 5 images.}
	\label{fig:visualization appendix}
\end{figure*}

The visualization results are in Figure~\ref{fig:visualization appendix}. From Figure~\ref{fig:visualization appendix}, we can have these findings:

\squishlist
    \item WSS can respond to part of the background classes, e.g., the first image on the left and the second image on the right. In these two images, WSS labels part of the tree as foreground while WSFL does not label them as foreground. Also, WSS labels parts of the shadow as salient objects in the first image on the right. 
    \item WSS can label only discriminative parts of the object in many images, like the second image on the left and the third image on the right. In these images, WSS only labels some parts of the objects (like the head of the bird and the cat). However, WSFL can label the whole object in the image. 
    \item For multiple objects with the same category in a single image, WSS will only label one most salient object and ignore other objects, e.g., the fourth row in the image. In contrast, WSFL can label all objects with the same category in a single image.
    \item WSS can ignore small objects like the third image on the left. There is an airplane in the image. WSFL can recognize the whole plane while WSS nearly missed this airplane. Considering the above three issues of WSS, it shows that weakly supervised saliency methods like WSS cannot predict foreground objects well. It may have several reasons. One reason is the small amount of training data. With a small set of training data, WSS cannot learn the object concept very well. Also, directly learning from classification labels will have some issues, like only learns the discriminative part of objects. However, our WSFL is learned from pseudo generated boxes, which can overcome these issues.

    \item Since WSS is based on high-resolution segmentation models while WSFL is based on classification models and WSFL is trained on masks generated by boxes, WSS can have better boundary predictions than WSFL. Also, when the background is not cluttered or the input image is rather simple, like the last row of Figure~\ref{fig:visualization appendix}, WSS can have better results than WSFL.
\squishend

\end{document}